\documentclass[conference]{IEEEtran}

\IEEEoverridecommandlockouts
\usepackage{cite}
\usepackage{amsmath,amssymb,amsfonts}
\usepackage{graphicx}
\usepackage{textcomp}
\usepackage{caption}
\usepackage{amsmath}
\usepackage{algpseudocode}
\usepackage{balance}
\usepackage[table,xcdraw]{xcolor}
\usepackage[hyphens]{url}
\usepackage{siunitx}
\usepackage{academicons}
\usepackage{scalerel}  

\usepackage{xcolor}
\usepackage{pict2e}

\newsavebox{\ORCIDlogo}
\savebox{\ORCIDlogo}{%
\setlength{\unitlength}{\dimexpr 1em/256\relax}%
\begin{picture}(256,256)%
  \color[HTML]{A6CE39}\put(128,128){\circle*{256}}%
  \color{white}%
  \put(78.6,199.2){\circle*{20}}%
  \moveto(70.9,176,9)\lineto(86.3,176,9)\lineto(86.3,69.8)\lineto(70.9,69.8)%
  \closepath\fillpath%
  \moveto(108.9,176.9)\lineto(150.5,176.9)%
  \curveto(190.1,176.9)(207.5,148.6)(207.5 ,123.3)%
  \curveto(207.5,95,8)(186,69.7)(150.7,69.7)%
  \lineto(108.9,69.7)%
  \closepath\fillpath%
  \color[HTML]{A6CE39}%
  \moveto(124.3,83.6)\lineto(148.8,83.6)%
  \curveto(183.7,83.6)(191.7,110.1)(191.7,123.3)%
  \curveto(191.7,144.8)(178,163)(148,163)%
  \lineto(124.3,163)%
  \closepath\fillpath%
\end{picture}%
}

\newcommand\orcidicon[1]{\href{https://orcid.org/#1}{\usebox{\ORCIDlogo}}}

\usepackage{hyperref} 

\begin{document}

\title{Federated Learning Method for Preserving Privacy in Face Recognition System}

\author{\IEEEauthorblockN{Enoch Solomon}
\IEEEauthorblockA{\textit{Department of Computer Science} \\
Virginia State University \\
Richmond, Virginia \\
esolomon@vsu.edu
}

\and
\IEEEauthorblockN{Abraham Woubie}
\IEEEauthorblockA{\textit{Silo AI} \\
Helsinki, Finland\\
abraham.zewoudie@silo.ai}
}


\maketitle

\begin{abstract}

The state-of-the-art face recognition systems are typically trained on a single computer, utilizing extensive image datasets collected from various number of users. However, these datasets often contain sensitive personal information that users may hesitate to disclose. To address potential privacy concerns, we explore the application of federated learning, both with and without secure aggregators, in the context of both supervised and unsupervised face recognition systems. Federated learning facilitates the training of a shared model without necessitating the sharing of individual private data, achieving this by training models on decentralized edge devices housing the data. In our proposed system, each edge device independently trains its own model, which is subsequently transmitted either to a secure aggregator or directly to the central server. To introduce diverse data without the need for data transmission, we employ generative adversarial networks to generate imposter data at the edge. Following this, the secure aggregator or central server combines these individual models to construct a global model, which is then relayed back to the edge devices. Experimental findings based on the CelebA datasets reveal that employing federated learning in both supervised and unsupervised face recognition systems offers dual benefits. Firstly, it safeguards privacy since the original data remains on the edge devices. Secondly, the experimental results demonstrate that the aggregated model yields nearly identical performance compared to the individual models, particularly when the federated model does not utilize a secure aggregator. Hence, our results shed light on the practical challenges associated with privacy-preserving face image training, particularly in terms of the balance between privacy and accuracy.
\end{abstract}


\section{Introduction}
\label{section:Introduction}

Face recognition is the process of automatically identifying or verifying the identity of an individual by analyzing facial patterns \cite{wang2022survey}. This technology has become an integral component in various security and authentication systems, ranging from smartphone unlocking \cite{article_phone}  to airport security checks \cite{article_airport}. It encompasses two primary subfields: face identification and face verification. Face identification determines the identity of an individual, whereas face verification confirms or denies a claimed identity \cite{wang2022survey}. Ensuring accurate face recognition is integral for granting access to services, as permissions should only be accorded following correct identification or verification.

The rapid advancement of machine learning (ML) and the availability of facial datasets have significantly enhanced the accuracy and performance of face recognition systems. Typically, face recognition systems employ machine learning techniques to train deep neural networks using facial data samples. Data samples are commonly gathered on end-devices like smartphones, while the model training takes place on a computationally robust centralized server~\cite{chen2019deep}. This setup raises two major concerns. First, since the model is trained on user face data, it is crucial to prevent unauthorized access or data breaches to protect user privacy. Second, such systems involve a heavy data transmission phase, which can place significant stress on the communication infrastructure. Federated learning offers a solution to both issues. Rather than sending the raw sensitive data to the central server for training the centralized model, federated learning advocates for a distributed training approach \cite{dean2012large}. In this setup, each device maintains its own instance of the model and trains it using its local data \cite{mcmahan2017communication}. After this local training, only the model updates are transmitted to the central server. The server then aggregates  these updates and applies them to the global model \cite{bonawitz2019towards,kairouz2019advances}. This approach ensures that sensitive facial data remains local, strengthening privacy measures and minimizing data transfers. The most common aggregation strategy is federated averaging, which aggregates the updates using a weighted average\cite{mcmahan2017communication}. 

Mobile phones and smart devices are examples of the modern distributed networks that generate huge amounts of data each day~\cite{li2020federated}. As these devices become more powerful and concerns about data privacy grow, federated learning has emerged as a notable solution to keep data on the device and shift the network's focus to the edge~\cite{li2020federated}. 
Various companies have adopted federated learning~\cite{bonawitz2019towards, wa2020federated},  highlighting its importance in applications that need privacy, especially when training data is spread across devices~\cite{huang2020federated,hard2018federated,yang2019federated,zhao2020mobile}. The increasing demand for federated learning across various applications has led to the development of numerous tools, including TensorFlow Federated~\cite{FedTensorFlow}, Federated AI Technology Enabler~\cite{FedAI}, Leaf~\cite{Leaf}, and PaddleFL~\cite{Paddle}. While privacy-preserving data studies have been of interest since the 1970s, it is only in recent times that they are being extensively employed at a large scale~\cite{erlingsson2014rappor}. For instance, Google uses federated learning in  Gboard~\cite{hard2018federated} and Android messages~\cite{AndroidFL}, and Apple has incorporated it in iOS 13~\cite{AppleFL}  for features like “Hey Siri”~\cite{SiriFL}.

As mentioned previously, privacy concerns are considered as one of the major challenges in face and speaker recognition systems~\cite{rahulamathavan2018privacy,woubie2021federated,woubie2021federatedsymp,solomon2023face,solomon2022uface,solomon2023fass,solomon2023hdlhc,solomon2023unsupervised,solomon2023autoencoder,woubie2023image,solomon2023deep} as these systems usually involve the complete sharing of facial data, which can bring threatening consequences to people’s privacy.  Federated learning emerges as a promising approach to address these concerns. Unlike conventional methods that require raw data to be sent to a central server for processing, federated learning enables model training directly on the user's device, ensuring that sensitive facial data remains local. This decentralized approach not only enhances privacy but also reduces the need for data transmission, thereby saving bandwidth.


Thus, the main contribution of this work centers on the integration of federated learning techniques in the training of deep neural network-based face recognition classifiers, both supervised and unsupervised, with the primary aim of safeguarding user privacy. In the proposed system, each device independently trains its own model and subsequently transmits this local model to either a secure aggregator or directly to a central server. The secure aggregator, in turn, consolidates these local models originating from various devices, assembles a global model, and dispatches it to the central server. Alternatively, the central server may construct the global model directly, without intermediary interaction with the secure aggregator. Ultimately, the central server redistributes the global model to all participating devices.

The proposed system facilitates the training of a face recognition model grounded in a deep neural network. It accomplishes this by utilizing data stored exclusively on the respective devices, guaranteeing that this data never exits the confines of those devices. The cloud-based component of the system employs federated averaging to merge these local models, thereby forming a global model that is subsequently relayed back to the devices for inference. The implementation of secure aggregation ensures that, at a global level, individual updates from the devices remain completely confidential and inscrutable. As the edge devices solely transmit model updates, no raw data ever departs from the edge. Consequently, the aggregator only has access to a model trained for the purpose of identifying a local user, preserving the privacy of all other information pertaining to face image at the edge.

A second innovation lies in the deployment of a generative adversarial network (GAN) to produce counterfeit data directly on edge devices. Employing a GAN eliminates the necessity of transmitting counterfeit data to the edge or accumulating such data at the edge itself. Transmitting counterfeit data could place a substantial strain on available bandwidth and, more crucially, expose potential vulnerabilities by revealing distinct information about the local user. Conversely, collecting counterfeit data at the edge could prove unfeasible.

The potential applications of the proposed federated learning systems are diverse and could encompass tasks like smartphone-based learning. By collaboratively learning facial characteristics from a multitude of mobile devices, a shared statistical model can be developed to effectively identify individuals. Nevertheless, users might be hesitant to relinquish their data to a central server, driven by concerns about safeguarding their personal privacy. As a solution, federated learning can be employed to train a centralized, user-independent model without the need to expose or share private data.

In the context of smartphone-based learning, a collective approach to learning face image characteristics from a substantial pool of mobile and similar devices enables the development of a unified statistical model for user identification. However, users may understandably harbor reservations about transferring their data to a central server, driven by privacy concerns. As a solution, federated learning can be employed to train a central, user-independent model without compromising the confidentiality of their private data.

In the context of learning across organizations, entities like universities can be likened to remote devices, each housing a wealth of student data. Nevertheless, universities are typically bound by stringent privacy regulations and practices, and any data leakage could lead to legal, administrative, or ethical complications. Federated learning offers a viable solution, allowing for confidential learning to take place across diverse devices and organizations while safeguarding the sensitive data of these institutions.

Our experiments conducted on the CelebA datasets reveal that federated learning brings notable advantages to both supervised and unsupervised facial recognition systems. This is achieved by avoiding the transmission of sensitive user data to central servers, while still delivering promising results when compared to individual local models. Consequently, the experimental outcomes provide a quantitative understanding of the challenges associated with the practical application of privacy-preserving training for facial recognition. These challenges are particularly evident in the trade-off between privacy and accuracy.


The remainder of this paper is structured as follows. Section~\ref{section:proposed_system} provides a detailed description of the proposed system's architecture. The experimental results are outlined in Section~\ref{section:experiment}, while Section~\ref{section:conclusion} presents the conclusions drawn from the work.

\section{Related Work}
Various methodologies have been proposed to enhance the privacy and security of face recognition systems.

\subsection{Privacy-Preserving Face Recognition}

Various methods have been explored to safeguard facial data.  Instead of using real images of individuals' faces, the authors of \cite{sFace} propose to generate synthetic images by training a class-conditional GAN. The synthetic data generator was trained on the original face dataset and the identities of the individuals as class labels. The authors then generate the synthetic dataset to train the face recognition model. PriFace \cite{zhao2023priface}  is another method for privacy-preserving face recognition. PriFace uses locality-sensitive hashing to add randomness to facial data, preventing potential misuse or reconstruction of the images. Further, the work in \cite{10026880} uses the Householder matrix to protect both model and facial data. This method combines additive and multiplicative perturbations, ensuring efficient user-side computations. For smart home settings, the authors of  \cite{10.1145/3472634.3472661}  propose to protect the face feature data of the users using a face recognition approach that combines random matrix and BLS short signature with FaceNet.  The work in \cite{GUO2019320}  proposes to protect the privacy of the faces by encrypting them through affine transformation, which consists of permutation, diffusion and shift transformations. Another scheme presented in \cite{Wang_Liu_Luo_Yang_Wang_2022} performs privacy-preserving face recognition scheme in the frequency domain. This scheme integrates an analysis network that gathers components with the same frequency from different blocks and  a fast masking method to further secure the remaining frequency components. We can also highlight the work of \cite{yang2022design} in which the normalized face feature vectors are encrypted using the CKKS algorithm from the SEAL library. To save computation costs that comes with encryption of query face images, \cite{10.1145/3448414} proposes to match an encrypted face query against clustered faces in the repository through a novel multi-matching scheme.

Other studies use local differential privacy to ensure that individual data points cannot be reverse-engineered or identified. The work of \cite{xie2023privacy} proposes a general privacy protection framework for edge-based face recognition systems. This is done through a local differential privacy algorithm based on the proportion difference of feature information. Furthermore, identity authentication and hash technology are used to ensure the legitimacy of the terminal device and the integrity of the face image in the data acquisition phase. 
The authors of  \cite{CHAMIKARA2020101951} introduce a new privacy-preserving face recognition protocol referred to as Privacy using EigEnface
Perturbation (PEEP). This protocol uses local differential privacy to apply perturbation to Eigenfaces. Only the perturbed data is stored in third-party servers, and a standard Eigenface recognition algorithm is run on this data.

\subsection{Federated Learning for Face Recognition}
Multiple methods use federated learning to ensure privacy-preserving  face recognition.  The authors of \cite{meng2022improving} introduce PrivacyFace, which leverages privacy-agnostic clusters during model training. These clusters are indifferent to privacy concerns (i.e., the data in these clusters do not reveal sensitive personal information).  PrivacyFace consists of two main components: the Differently Private Local Clustering (DPLC) algorithm, which derives privacy-independent group features, and a consensus-aware face recognition loss that refines the global feature space distribution using these desensitized group features. FedFace  \cite{aggarwal2021fedface} presents a federated learning framework that learns from face images across multiple clients without sharing the images with other clients or a central host. Each client, typically a mobile device, contains face images of only its owner. Face Presentation Attack Detection (FedPAD) \cite{shao2020federated}  aims to develop generalized fPAD models while ensuring data privacy. Each data owner trains a local fPAD model, and a server aggregates these models without accessing individual private data. Once the global model is refined, it's used for fPAD inference. FedFR \cite{Liu_Wang_Chien_Lai_2022} is a federated learning-based framework for privacy-aware generic face representation. The framework optimizes personalized models for clients using the Decoupled Feature Customization module, improving both the global model for face representation and the personalized user model.

\section{Proposed System}
\label{section:proposed_system}

\label{section:proposed_system}

The challenge of federated learning revolves around the task of constructing a unified global statistical model using data distributed across a limited number to possibly millions of remote devices. More specifically, the primary objective commonly pursued in federated learning is the minimization of the following objective function:

\begin{equation}
    \min _{w} F(w), \text { where } F(w):=\sum_{k=1}^{m} p_{k} F_{k}(w),
\end{equation}
where $m$ represents the total number of devices, $F_k$ denotes the local objective function for the $k$th device, and $p_k$ signifies the relative impact of each device with $p_{k} \geq 0$ and $\sum_{k=1}^{m} p_{k}=1$.

Federated learning empowers the distributed training of face recognition models, accommodating a diverse range of client devices. As illustrated in Fig. 1, the envisioned federated learning system for face recognition functions across three key locations: edge devices, a secure aggregator, and a central main server. These edge devices encompass a variety of hardware, including mobile phones, laptops, and similar devices. In contrast, the aggregator and main server typically operate as cloud-based services.

Fig. 1 depicts the training of a central model using a distributed dataset. Here, a multitude of nodes, which could represent user devices, possess subsets of data with varying sizes. At the device level, each node computes a model update, which is subsequently conveyed to a central server. During each training iteration, a substantial volume of these updates or gradients is amalgamated at the central server. The central server then derives a global update for the central model by computing the average of these individual local updates.

Note that the architecture of the proposed system remains consistent for both supervised and unsupervised face recognition systems. The key distinction lies in the utilization of labels for training in supervised systems, while unsupervised systems do not rely on labels for training individual face recognition models.

While it is feasible to train individual face recognition models in the supervised system using only client images of a given person on a specific device, our preference is to enhance model robustness and improve the ability to distinguish impostor images. To achieve this, as depicted in Fig. 1, we employ two distinct methods for generating impostor image data for each individual on the edge device:

\begin{itemize}
    \item In the first method, we randomly select the image of other persons from the CelebA dataset as impostor image data for a given person. 
    
    \item In the second method, we train a GAN model to generate impostor image data as it is not always easy to find image data  of different persons in edge devices. Thus, we use the work of~\cite{donahue2019wavegan} to train a GAN model on the CelebA dataset. Once the impostor images are generated using the trained GAN model, they are combined with client image data to train an individual face recognition model on a specific edge device. 
\end{itemize}

\begin{figure}[t!]
\centering
\includegraphics[width=\columnwidth]{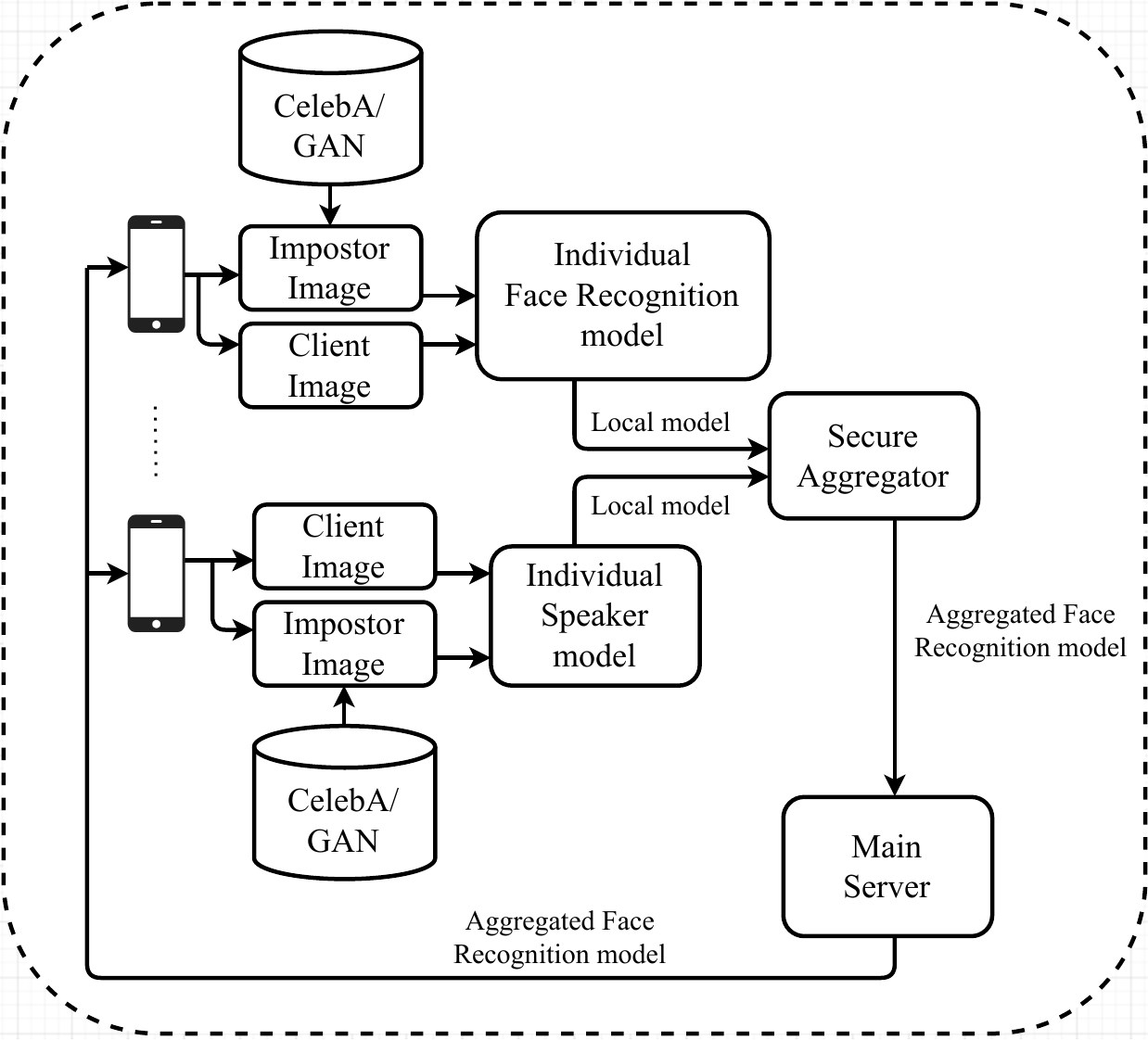}
\caption{The proposed face recognition system incorporates federated learning. Through the implementation of a secure aggregator, we empower a collective of inherently untrusting devices to collaborate and calculate an aggregate value without disclosing their individual private data.}\label{fig:proposed_federated}
\vspace{-0.6cm}
\end{figure}

The proposed system employs distributed gradient descent to train a deep neural network across training data residing on user-held devices, with the aim of analyzing the impact of a secure aggregator. In the system that incorporates a secure aggregator, the process unfolds as follows:

\begin{enumerate}
 \item Local Training: Initially, an individual model is trained locally on each user's device.
 \item Model Transmission to Secure Aggregator: Subsequently, each user's device transmits its locally trained model to the secure aggregator.
 \item Global Model Creation: The secure aggregator aggregates these individual models to construct a global model.
 \item Aggregated Model Transmission: The aggregated model is then sent to the central main server.
 \item Distribution to Devices: Finally, the main server redistributes the global model to each individual device.
\end{enumerate}

In contrast, in the system where a secure aggregator is not utilized, the workflow proceeds as follows:

\begin{enumerate}
 \item Local Training: Each device independently conducts local training to create an individual model.
 \item Model Transmission to Main Server: These individual models are directly transmitted to the central main server.
 \item Global Model Creation: The main server combines these individual models to form a global model.
 \item Aggregated Model Transmission: The global model is sent back to each individual device for further use and updates.
\end{enumerate}

This dual approach allows for a comparative analysis of the system's performance with and without the incorporation of a secure aggregator.

Privacy serves as a significant driving force behind the adoption of federated learning applications. These systems are designed to safeguard user data by prioritizing the sharing of model updates, such as gradient information, rather than the raw and potentially sensitive data itself. This innovative approach to collaborative machine learning not only enhances data privacy but also enables the collective training of robust and accurate models without exposing individual user information to undue risks or breaches~\cite{carlini2018secret,wainwright2012privacy,dwork2014algorithmic}. While federated learning mitigates some privacy risks by not directly sharing raw data, it's important to recognize that sending model updates during the training process can still pose potential privacy challenges~\cite{mcmahan2017learning}. While recent advancements in federated learning have made strides in enhancing privacy through tools like secure multiparty computation (SMC) or differential privacy (DP), these approaches have trade-offs between privacy and model performance. The secure aggregator belongs to the class of secure multi-party computation algorithms, where a set of inherently distrustful devices denoted as $d \in U$ individually possess private values $x_u$. These devices collaborate to calculate an aggregate value, such as the sum $\sum_{u \in U}x_u$, while ensuring that no device discloses any information about its private value to others, except what can be inferred from the resulting aggregate value.

The proposed system aims to uphold the privacy of federated learning by incorporating the use of secure multiparty computation (SMC) techniques.~\cite{bonawitz2017practical, ghazi2019scalable}. The adoption of secure multiparty computation serves to safeguard individual model updates, ensuring their privacy and confidentiality. The central server is unable to observe individual local updates; it can only access the aggregated results at each round.

The proposed work employs the classical federated learning average (FedAvg)\cite{mcmahan2017communication}. The process involves local optimization executed on participating clients and a subsequent server step to update the global model. Notably, Algorithm 1 illustrates that devices communicate only the updated weights rather than face image data, preserving the security and privacy of the user's facial information locally.

Addressing the transfer of a substantial volume of updated model parameters from users to a server, which is often restricted in throughput ~\cite{brendan2016communication,kairouz2019advances,li2020federated,chen2020joint}, poses a significant obstacle in federated learning. This difficulty can be addressed through strategies such as minimizing the number of participating users, achieved through the implementation of scheduling policies ~\cite{yang2019scheduling,amiri2020update}.

\section{Experiments}\label{section:experiment}

\subsection{Dataset} 
CelebA (Celebrities Attributes Dataset)  \cite{liu2015deep},  is a popular dataset in the field of computer vision and machine learning. It was created by researchers at the Chinese University of Hong Kong and is often used for various facial recognition and image analysis tasks. CelebA is known for its large collection of celebrity images and the annotations associated with them. CelebA contains more than 200,000 celebrity images. These images cover a wide range of celebrities from different backgrounds and professions.  Each image in the CelebA dataset is annotated with a set of 40 binary attributes. These attributes include characteristics like "smiling," "wearing glasses," "wearing a hat," and so on. These annotations are valuable for tasks like facial attribute prediction and facial attribute manipulation. In addition to attribute annotations, CelebA also provides identity labels for the celebrities in the dataset. This can be useful for tasks involving face recognition. The images in CelebA showcase a wide variety of poses, expressions, lighting conditions, and backgrounds, making it suitable for a broad range of computer vision tasks. The dataset is typically split into training, validation, and test sets to facilitate model training and evaluation.

\subsection{Experimental Setup}

The system architectures are:
\begin{itemize}
    \item The CNN architecture utilized in our work closely mirrors VGG-M \cite{chatfield2014return}, a widely adopted architecture for image classification and speech technology applications \cite{chung2016out}. Furthermore, we incorporate a max-pooling layer with dimensions of 2 by 2, along with batch normalization and dropout layers.
    
    The supervised system has been implemented using the Keras deep learning library~\cite{chollet2017keras} to train the model. The network is trained on Titan X GPUs for 100 epochs or until the validation error stops decreasing, whichever is sooner, using a batch-size of 64. We use SGD with momentum (0.9), weight decay (\num[{scientific-notation = true}]{5E-4}) and a logarithmically decaying learning rate (initialised to $10^{-2}$ and decaying to $10^{-8}$).
    
\begin{table}[t]
\caption{The architecture employed for the supervised face verification system..} 
\centering 
\begin{tabular}{c c c c} 
\hline\hline 
\textbf{Layer} & \textbf{Kernel} & \textbf{Filters} & \textbf{Output size} \\ [0.5ex] 
\hline 
Conv-1 & 3 X 3 & 64 & 350 X 80 X 64 \\ 
Conv-2 & 3 X 3 & 128 & 175 X 40 X 128 \\
Conv-3 & 3 X 3 & 256 & 87 X 20 X 256 \\
fc-1 & - & 1000 & - \\
fc-2 & - & 400 & - \\
fc-3 & - & 1 & - \\
\hline 
\end{tabular}
\label{table:architecture}
\vspace{-0.25cm}
\end{table}
    
    \item An autoencoder is employed to train the unsupervised system with the primary objective of enabling the network to acquire a representation of person-specific facial data. The CNN component, identified as the encoder, is optimized to learn a sophisticated representation of the provided facial image, while the decoder component is fine-tuned to reconstruct the encoder's output into the corresponding facial image. Following the training phase, the decoder component is discarded, and the already learned encoding representation is repurposed for the face verification task. The unsupervised system does not utilize impostor data since its primary focus is on acquiring a compact vector representation of distinct individual faces.

\end{itemize}

\begin{figure}[h]
\centering
\includegraphics[width=\columnwidth]{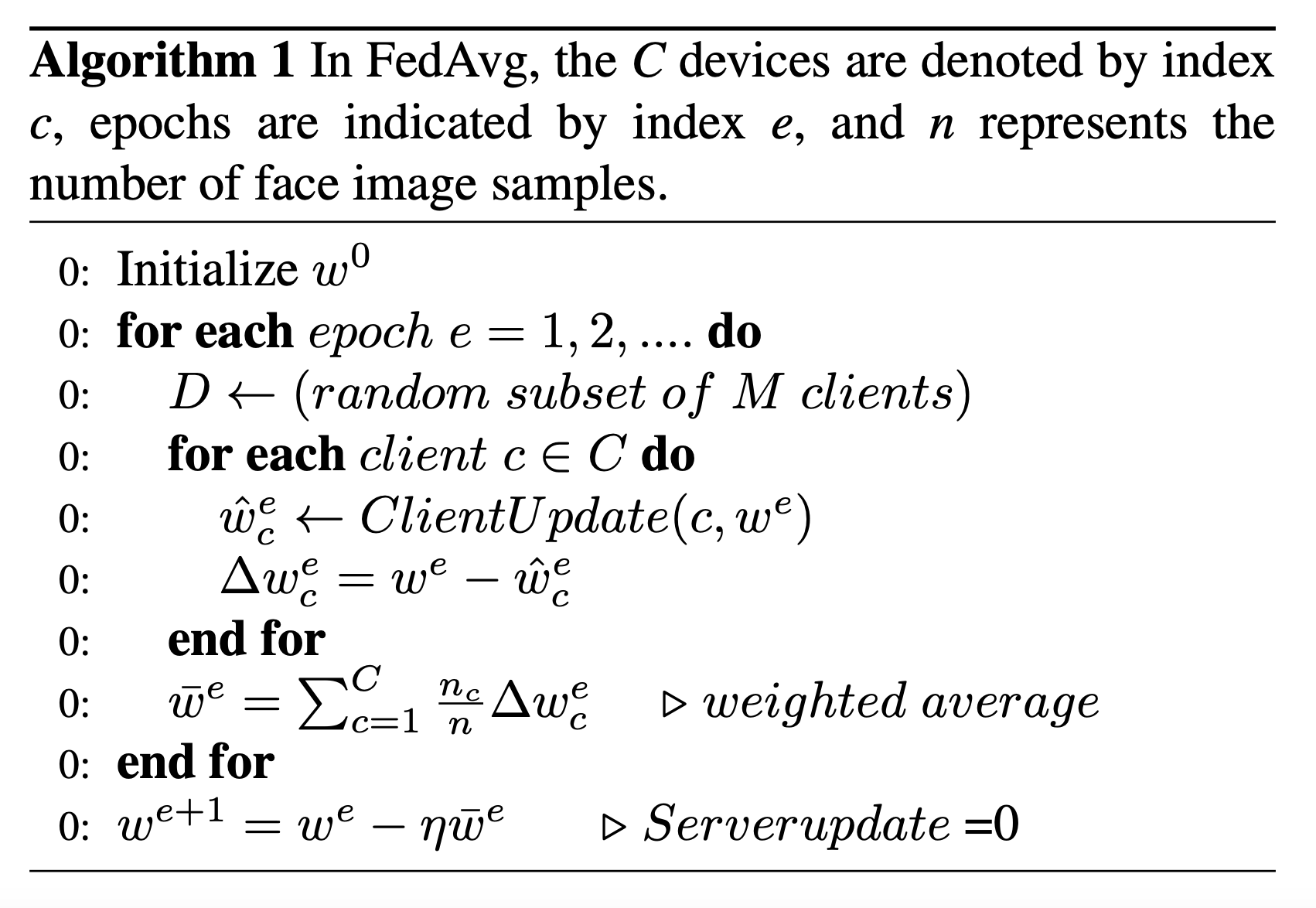}
\label{alg:1}
\vspace{-0.5cm}
\end{figure}

The proposed face verification system has been carried out on the following databases namely CelebA ~\cite{liu2015deep}. We randomly selected 1000 persons' face images. We allocated 90\% of each person's face images for training an individual, face-dependent model, while the remaining 10\% was reserved for evaluation. For instance, if a given person had 100 face images in the development set from the database, 90 images were utilized for training the individual face model, and the remaining 10 images were used for evaluation. Additionally, impostor data was introduced into the test set for comprehensive assessment.

Initially, our intention was to train individual face models exclusively using the authentic client face data for each person on every device. However, due to the limited number of files for each individual in the dataset—most individuals having fewer than 100 face images—this approach resulted in an overfitting problem. To address this, we modified our strategy and trained individual face models by incorporating both the true face of the individual and the face images of other individuals as impostor face data.

We adopted two distinct methods to generate impostor face images for each individual device. In the first method, we selected face images of other individuals from each dataset as impostor face images, with 100 samples chosen for each individual on a given device. For the second method, impostor data was created using a GAN model, leveraging the approach outlined in ~\cite{donahue2019wavegan} to train the GAN model on each dataset. Similar to the first method, we generated 100 impostor face images for each individual device.

The primary challenge in training the GAN model to generate impostor face images lies in its time-consuming training phase. The computational cost of training the GAN model for 50 hours on the CelebA dataset with a Quadro P2000 GPU amounts to 3.5 hours. However, once the GAN model is trained, the extraction of impostor face image samples on edge devices becomes significantly faster. It's important to note that the training of the GAN model is a one-time task.

The proposed system's performance is assessed using the Equal Error Rate (EER), a metric that measures the point at which the rates of acceptance and rejection errors are equal.

\subsection{Experimental Results} 

As it is mentioned in Section~\ref{section:proposed_system}, we have analyzed the impact of federated learning both for supervised and unsupervised face verification systems with and without using the secure aggregator. Thus, the experimental results of the supervised and unsupervised systems with and without using the secure aggregator are described below.


\subsubsection{Supervised Systems without Secure Aggregator}

Fig.~\ref{fig:supervised_system_withou_SA} illustrates the comparative performance of both individual and aggregated models within the supervised system, with and without the utilization of GAN. Notably, the distinction between Fig.~\ref{fig:supervised_system_withou_SA} (a) and Fig.~\ref{fig:supervised_system_withou_SA} (b) lies in the method of generating impostor face image samples. In Fig.~\ref{fig:supervised_system_withou_SA} (a), impostor face image samples are created by selecting face images of different individuals (i.e., extracting face images from CelebA to serve as impostors for a given face image). Conversely, in Fig.~\ref{fig:supervised_system_withou_SA} (b), the GAN model is employed to generate the impostor face images.


The primary distinction between the individual and aggregated face image models lies in their training approach. For the individual model, a dedicated face image model is initially trained for each specific face image, utilizing the corresponding individual's face image data. Subsequently, the face image samples are assessed using this personalized face image model. In this work, the individual model serves as the baseline system, wherein 1000 individual face image models are trained using face image samples from 1000 devices.

In contrast, the aggregated model employs a collaborative approach. Each of the 1000 devices transmits their parameters to a secure aggregator. The aggregator computes the average of these parameters, establishing them as its updated weight parameters, and then redistributes them to the 1000 devices. Consequently, this collaborative model is referred to as the aggregated (federated) face image model.

\begin{figure}[t]
    \centering
    {{\includegraphics[width=8cm]{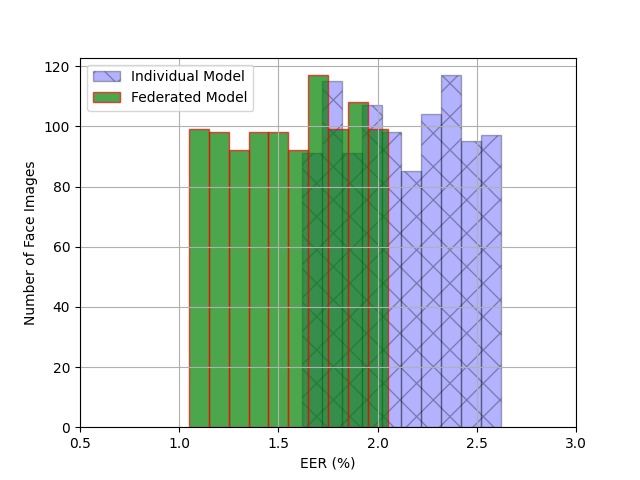} }}%
    (a)\;Impostors selected from CelebA dataset.
    \qquad
    {{\includegraphics[width=8cm]{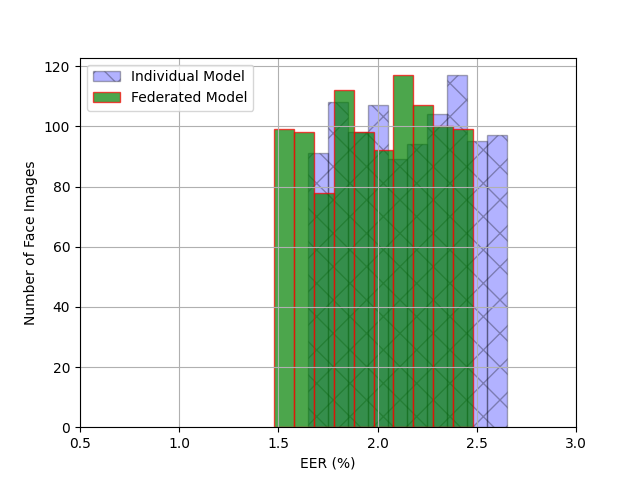} }}%
    (b)\;Impostors generated using GAN.
    \caption{Histograms depicting the Equal Error Rate (EER) across 1000 devices are presented for the comparison between individual and federated models in the supervised systems. Notably, this evaluation focuses on models that do not utilize a secure aggregator (SA).}%
    \label{fig:supervised_system_withou_SA}
\end{figure}

In Fig.~\ref{fig:supervised_system_withou_SA}, it is evident that when each of the 1000 devices employs its own individual model, the Equal Error Rate (EER) surpasses 1.98. However, with the utilization of federated/aggregated models, irrespective of the impostor generation method, a majority of the devices exhibit EER values below 1.98. Specifically, employing the first method, which uses the face images of other individuals as impostor data, around 324 devices yield an EER below 1.98. Meanwhile, in the second aggregation method involving GAN-generated impostor face images, a similar number of devices achieve an EER below 1.98.

In both Fig.\ref{fig:supervised_system_withou_SA} (a) and Fig.\ref{fig:supervised_system_withou_SA} (b), it is noticeable that the aggregated face image model consistently outperforms the individual model in terms of average EER, regardless of whether GAN or other persons' face images are used as impostor data. These figures also indicate that the two aggregated methods yield nearly identical average EER values. This suggests the feasibility of employing GAN for on-device generation of impostor face images, eliminating the need to transfer impostor data from external sources to edge devices.

Table~\ref{table:results_supervised} presents a comprehensive overview of the results. The average EER of individual models across the 1000 devices/face images in the supervised face image verification system stands at 2.11. This average EER serves as the baseline system, calculated by utilizing data trained specifically for each face image/device.

Additionally, Table~\ref{table:results_supervised} highlights that the average EER for the 1000 devices under the federated model, without employing a secure aggregator and GAN, is 1.55. This represents a noteworthy 26.5\% relative improvement compared to the baseline systems. Similarly, leveraging GAN to generate impostor data during the training of face image models yields superior EER results compared to the baseline system. The table illustrates that the federated model utilizing GAN-generated data achieves an average EER of 1.98, reflecting a 6

\subsubsection{Supervised Systems using Secure Aggregator}

Table~\ref{table:results_supervised} reveals that the average Equal Error Rate (EER) of the 1000 devices within the federated model of the supervised system, incorporating both a secure aggregator and impostor face images from the CelebA dataset, is 2.61. Similarly, the average EER for the 1000 devices in the federated model of the same system, employing the GAN technique to generate impostor face images, is 2.73. These findings suggest that, regardless of the impostor generation method, the inclusion of a secure aggregator leads to inferior results compared to both individual systems and federated systems that do not involve a secure aggregator.

The decline in EER when employing a secure aggregator in the federated system can be attributed to the trade-off between privacy enhancement and model performance or system efficiency. While recent approaches aim to bolster the privacy of federated learning through secure aggregation, this often comes at the expense of reduced model performance or overall system efficiency. Consequently, it becomes essential to weigh the privacy aspect alongside the EER values. Nevertheless, the results from systems incorporating secure aggregators remain acceptable despite the observed trade-offs.

In Figure~\ref{fig:box_plot_supervised}, the distribution of Equal Error Rate (EER) among the 1000 devices is presented for both individual and aggregated models within the supervised system, with and without the inclusion of a secure aggregator. The figure also highlights the influence of using Generative Adversarial Networks (GAN) for impostor face image generation. The depicted elements include the minimum, lower quartile, median, upper quartile, and maximum EER values.

\begin{figure}[h]
    \centering
    {{\includegraphics[width=8cm]{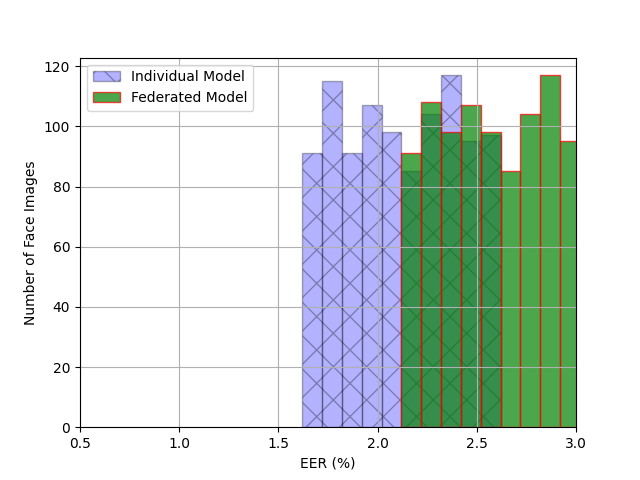} }}%
    (a)\;Impostors selected from CelebA dataset.
    \qquad
    {{\includegraphics[width=8cm]{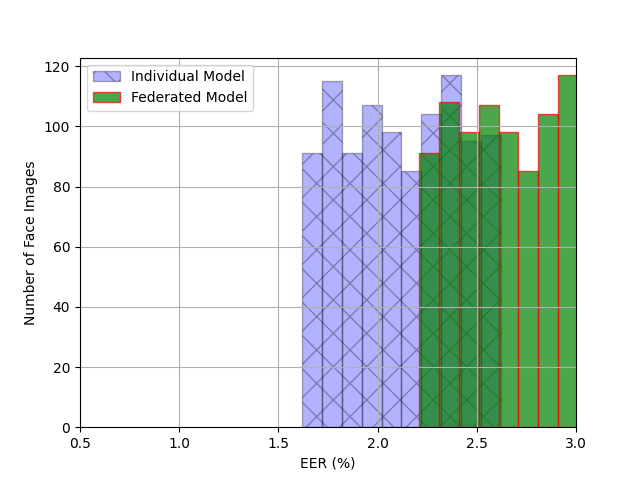} }}%
    (b)\;Impostors generated using GAN.
    \caption{Histograms illustrating the Equal Error Rate (EER) distribution across 1000 devices are provided for a comparison between individual and federated models in the supervised system. This analysis specifically considers models that incorporate a secure aggregator (SA).}
    \label{fig:supervised_system_withou_SA}
\end{figure}

As evident from the figure, aggregated models, particularly those not incorporating a secure aggregator, consistently outperform individual models in terms of average EER, regardless of the impostor generation method employed. The visual representation of EER distribution provides a clear indication of the superior performance of aggregated models, reinforcing the efficacy of collaborative approaches in contrast to individual models within the supervised system.

\begin{figure}[t!]
\centering
\includegraphics[width=\columnwidth]{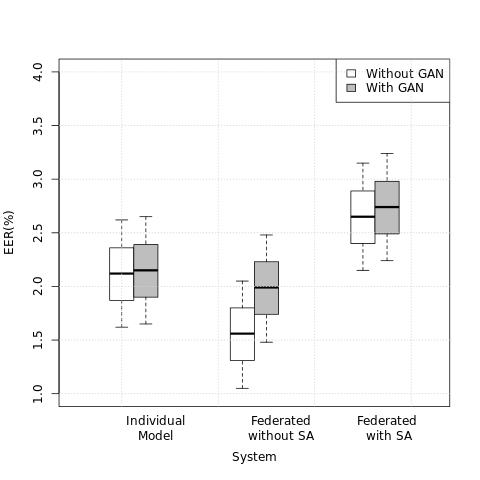}
\caption{The box plot depicts the distribution of Equal Error Rates (EER) for both supervised individual and federated models across 1000 devices. The analysis considers scenarios both with and without using a Secure Aggregator (SA). Additionally, the influence of impostor selections, with and without the incorporation of Generative Adversarial Network (GAN), is highlighted.}
\label{fig:box_plot_supervised}
\vspace{-0.5cm}
\end{figure}

\begin{table}[t!]
\caption{The table provides a comparison of Equal Error Rates (EER) for supervised face verification systems, considering both individual and federated approaches, with and without the use of a Secure Aggregator (SA). The inclusion of GAN-generated data for impostor face images is also accounted for in the comparison.} 
\centering 
\begin{tabular}{c c c} 
\hline\hline 
\textbf{System} & \textbf{With SA} & \textbf{Without SA} \\ [0.5ex] 
\hline 
Individual Model & \multicolumn{2}{c}{2.11} \\ 
Federated Model without GAN & 2.61 & 1.55 \\
Federated Model with GAN & 2.73 & 1.98 \\
\hline 
\end{tabular}
\label{table:results_supervised}
\vspace{-0.25cm}
\end{table}

\subsubsection{Unsupervised Systems}

\begin{figure}[t]
    \centering
    {{\includegraphics[width=8cm]{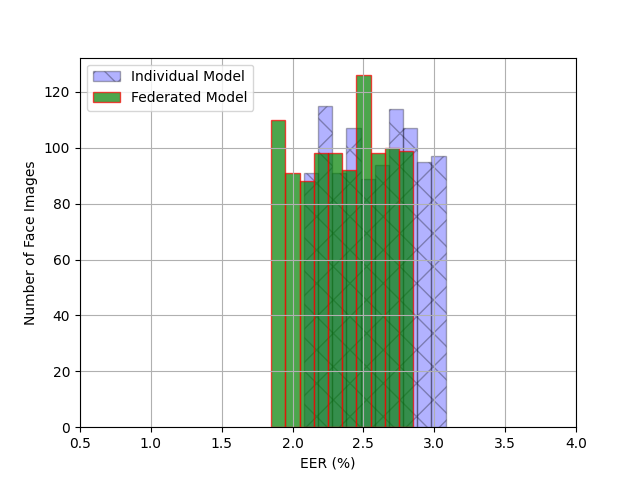} }}%
    (a)\;Without using a secure aggregator (SA).
    \qquad
    {{\includegraphics[width=8cm]{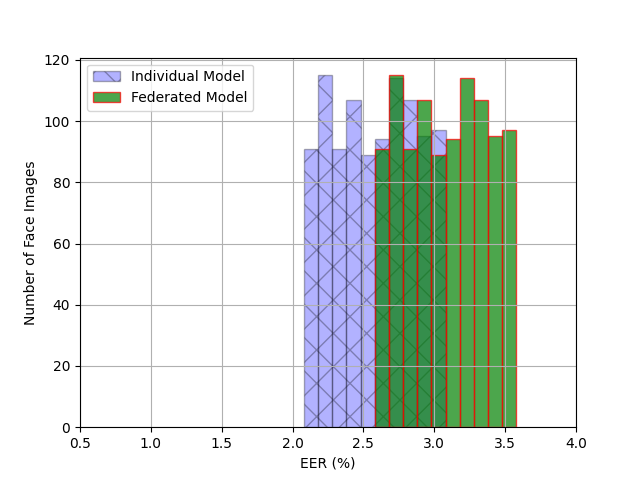} }}%
    (b)\;Using a secure aggregator.
    \caption{The Equal Error Rate (EER) across 1000 devices is reported for the comparison between the individual and federated models in the unsupervised system.}%
    \label{fig:unsupervised_system}
\end{figure}

In Fig.~\ref{fig:unsupervised_system}, the Equal Error Rates (EERs) are presented for both individual and aggregated face image models within the unsupervised system, with and without the implementation of a secure aggregator. As depicted, the EER for all 1000 devices exceeds 2.57 when each device utilizes its individual model. However, when the federated model does not employ a secure aggregator, as shown in Fig.~\ref{fig:unsupervised_system} (a), approximately 680 devices achieve an EER below 2.35.

This visual representation underscores a notable improvement in EER when transitioning from individual models to federated models without a secure aggregator in the unsupervised system. The collaborative approach appears to enhance the performance of the face image models, contributing to lower EER values for a significant portion of the devices.

In contrast to the results depicted in Fig.\ref{fig:unsupervised_system} (b), the use of a secure aggregator results in a deterioration of Equal Error Rates (EERs), leading to inferior results when compared to the individual models. This observation highlights a significant discrepancy in performance when incorporating a secure aggregator within the unsupervised system. The visual representation in Fig.~\ref{fig:unsupervised_system} (b) underscores the importance of carefully evaluating the influence of a secure aggregator on EER results, revealing a potential trade-off between privacy-enhancing measures and model performance in the context of the unsupervised system.

\begin{table}[t]
\caption{The Equal Error Rate (EER) is compared between the unsupervised face verification systems, considering both individual and federated approaches, with and without the use of a Secure Aggregator (SA).} 
\centering 
\begin{tabular}{c c c} 
\hline\hline 
\textbf{System} & \textbf{With SA} & \textbf{Without SA} \\ [0.5ex] 
\hline 
Individual Model & \multicolumn{2}{c}{2.57} \\ 
Federated Model & 3.07 & 2.35 \\
\hline 
\end{tabular}
\label{table:results_unsupervised}
\end{table}


Table~\ref{table:results_unsupervised} displays the average Equal Error Rate (EER) of individual models across the 1000 persons within the unsupervised face verification system, amounting to 2.57. In contrast, the table reveals that the average EER for the 1000 devices in the federated model, under the same system but without a secure aggregator, is 2.35. This signifies an 8.56\% relative improvement in EER compared to the baseline unsupervised system. However, it is noteworthy that the inclusion of a secure aggregator in the federated model results in a worse outcome compared to the baseline system, as indicated in the table.

\begin{figure}[h]
\centering
\includegraphics[width=\columnwidth]{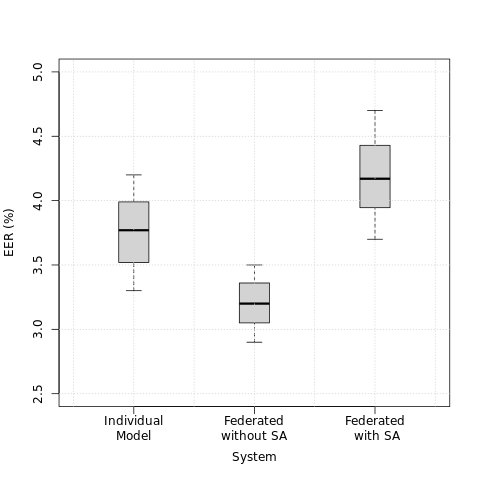}
\caption{A box plot is presented to illustrate the distribution of Equal Error Rates (EER) for unsupervised individual and federated models across 1000 devices. The analysis encompasses scenarios both with and without the utilization of a Secure Aggregator (SA).}
\label{fig:box_plot_unsupervised}
\vspace{-0.5cm}
\end{figure}

\subsection{Discussions}

The results presented in Table~\ref{table:results_supervised} and~\ref{table:results_unsupervised} consistently highlight that, regardless of the face verification system (supervised or unsupervised), the introduction of a secure aggregator tends to reduce the performance of the federated model. Conversely, when the federated model operates without a secure aggregator, EER results improve in comparison to the individual models. Despite the slight deterioration caused by the secure aggregator, it's essential to consider that its inclusion ensures the privacy of the data.

It's worth noting that, while the EER results experience a slight decline with the secure aggregator, the overall performance remains satisfactory. The compromise in EER is balanced by the privacy-preserving benefits offered by the secure aggregator. The EER results, even with the use of a secure aggregator, are acceptable, emphasizing the trade-off between privacy protection and model performance.

In addition to the individual and federated models, another experiment was conducted by pooling all face image samples from the 1000 persons and training a single generic face image model on a single computer. The results demonstrate that the average EER of the global model on the supervised and unsupervised face verification systems is 1.2\% and 2.2\%, respectively. These results are comparable to the federated model's performance (see Fig. \ref{fig:box_plot_supervised} and \ref{fig:box_plot_unsupervised}). The federated model achieves similar EER values as the global model while preserving the privacy of face image data. This underscores the advantage of using federated learning for face recognition systems.

The work employs 1000 devices to compare the performance of individual versus federated models. Statistical analysis using Student's t-test supports the significance of the observed differences. The computed P-values for both comparisons, where the federated model selects impostor face images from CelebA (federated model 1) and where GAN is used for impostor face image generation (federated model 2), are both less than the standard significance level of 0.05. Thus, we reject the null hypothesis, affirming that the mean EER differences between individual and federated models are statistically significant.

Finally, the experiment considered the impact of updating local models more than once. The results indicate that updating local models more than once does not lead to an improvement in EER. This could be attributed to the similarity in training data among devices during each training phase. Although updating more frequently did not yield enhanced performance, this decision was driven by the need to maintain data privacy.

\section{Conclusions}
\label{section:conclusion}

In this work, we propose the adoption of federated learning as a safeguard for the privacy of facial image data residing on edge devices, applicable to both supervised and unsupervised face recognition systems. Our approach centers on decentralized training, eliminating the need for devices to transmit their raw image data to centralized servers. Instead, each user's data remains securely stored and processed solely on their respective edge device. Consequently, training occurs exclusively at the local level, with each device contributing updates to a central model. Subsequently, a secure aggregator consolidates these local models into a single federated model, which is then distributed via the main server back to the individual devices. Furthermore, our research delves into an analysis of the influence of the secure aggregator on the performance of face recognition systems.

Our proposed system offers two primary advantages. Firstly, as raw data remains confined to individual devices, the privacy of facial images is preserved. Secondly, experimental findings reveal that the federated model, devoid of a secure aggregator, achieves a superior average Equal Error Rate (EER) compared to individual models. However, when the federated model incorporates the secure aggregator, the aggregated model yields EER results that are slightly less favorable than those of individual models. Nonetheless, the EER results remain commendable, emphasizing the importance of weighing the trade-offs between privacy and performance.

Future research works should delve into refining aggregation techniques beyond simplistic averaging methods. Additionally, exploring the effects of scaling up the number of devices beyond the 1000 devices employed in this work holds promise for further enhancing the effectiveness of privacy-preserving face recognition systems.

\end{document}